\documentclass[sigconf]{acmart}

\copyrightyear{2022}
\acmYear{2022}
\setcopyright{rightsretained}
\acmConference[CIKM '22]{Proceedings of the 31st ACM International Conference on Information and Knowledge Management}{October 17--21, 2022}{Atlanta, GA, USA}
\acmBooktitle{Proceedings of the 31st ACM Int'l Conference on Information and Knowledge Management (CIKM '22), Oct. 17--21, 2022, Atlanta, GA, USA}
\acmDOI{10.1145/3511808.3557484}
\acmISBN{978-1-4503-9236-5/22/10}

\usepackage{etoolbox}
\makeatletter
\patchcmd{\maketitle}{\@copyrightpermission}{
   \begin{minipage}{0.3\columnwidth}
     \href{https://creativecommons.org/licenses/by/4.0/}{\includegraphics[width=0.90\textwidth]{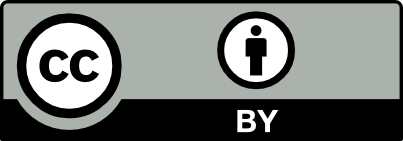}}
   \end{minipage}\hfill
   \begin{minipage}{0.7\columnwidth}
     \href{https://creativecommons.org/licenses/by/4.0/}{This work is licensed under a Creative Commons Attribution International 4.0 License.}
   \end{minipage}
  
   \vspace{5pt}
}{}{}

\makeatother

\usepackage[T1]{fontenc}
\usepackage{algpseudocode}
\usepackage[linesnumbered,ruled]{algorithm2e}
\usepackage{graphicx}
\usepackage{colortbl}
\usepackage{arydshln}
\usepackage{xcolor}
\usepackage{caption}
\usepackage{subcaption}
\usepackage{balance}
\usepackage{comment}
\usepackage{multirow}
\usepackage{multicol}
\usepackage{booktabs}
\usepackage{fontawesome}
\usepackage{hyperref}
\usepackage{makecell}

\usepackage{amssymb}
\usepackage{amsmath}
\usepackage{pifont}
\newcommand{\cmark}{\ding{51}}%
\newcommand{\xmark}{\ding{55}}%
\newcommand{\ha}[1]{\textcolor{red}{Hiba: #1}}

\newcommand{\uncommon}[1]{{\fontfamily{ppl}\selectfont \textsc{UnCommonSense}}}
\usepackage{tikz-cd} 
\usepackage{pgfplots}
\usetikzlibrary{patterns}
\usetikzlibrary{plotmarks}
\pgfplotsset{compat=1.11,
    /pgfplots/ybar legend/.style={
        /pgfplots/legend image code/.code={%
            \draw[##1,/tikz/.cd,bar width=5pt,yshift=-0.2em,bar shift=0pt]
            plot coordinates {(0cm,0.8em)};
        },
    },
}

\begin{document}

\title[UnCommonSense: Informative Negative Knowledge about Everyday Concepts]{UnCommonSense:\\ Informative Negative Knowledge about Everyday Concepts}

\author{Hiba Arnaout}
\email{harnaout@mpi-inf.mpg.de}
\affiliation{%
  \institution{Max Planck Institute for Informatics}
  \city{Saarbrücken}
  \country{Germany}
}

\author{Simon Razniewski}
\email{srazniew@mpi-inf.mpg.de}
\affiliation{%
  \institution{Max Planck Institute for Informatics}
  \city{Saarbrücken}
  \country{Germany}
}

\author{Gerhard Weikum}
\email{weikum@mpi-inf.mpg.de}
\affiliation{%
  \institution{Max Planck Institute for Informatics}
  \city{Saarbrücken}
  \country{Germany}
}

\author{Jeff Z. Pan}
\email{j.z.pan@ed.ac.uk}
\affiliation{%
  \institution{The University of Edinburgh}
  \city{Edinburgh}
  \country{United Kingdom}
}

\renewcommand{\shortauthors}{Hiba Arnaout, Simon Razniewski, Gerhard Weikum, \& Jeff Z. Pan}

\begin{abstract}
 Commonsense knowledge about everyday concepts is an important asset for AI applications, such as question answering and chatbots. Recently, we have seen an increasing interest in the construction of structured 
commonsense knowledge bases (CSKBs).
An important part of 
human commonsense
is about properties that do \textit{not} apply to concepts, yet existing CSKBs 
only store positive statements. 
Moreover, since CSKBs operate under the open-world assumption, 
absent statements are considered to have unknown truth rather than being invalid.
This paper presents the \uncommon{} framework for materializing informative negative commonsense 
statements. 
Given a target concept, comparable concepts are identified in the CSKB, for which a local closed-world assumption 
is postulated.
This way, positive statements about comparable concepts that are absent for the target concept become seeds for negative statement candidates.
The large set of candidates is then scrutinized, pruned and
ranked by informativeness. Intrinsic and extrinsic evaluations show that our method significantly outperforms the state-of-the-art. A large dataset of informative negations is released as a resource for future research.
\end{abstract}


\maketitle

\section{Introduction}
\label{sec:intro}
\noindent
\textbf{Motivation. } Commonsense knowledge (CSK) is crucial for robust AI applications such as question answering and chatbots. The purpose is to enrich machine knowledge with properties about everyday concepts (e.g., gorilla, pancake, newspaper). Such statements are acquired, organized and stored in structured knowledge bases (KBs)~\cite{Pan2016,DBLP:series/synthesis/2021Hogan,DBLP:journals/ftdb/WeikumDRS21}. 
Large commonsense KBs (CSKBs) include ConceptNet~\cite{conceptnet}, WebChild~\cite{webchild}, 
ATOMIC~\cite{atomic}, 
TransOMCS~\cite{DBLP:conf/ijcai/ZhangKSR20},
and Ascent~\cite{ascent}. 
These projects are almost exclusively focused on positive statements such as \textit{gorillas are mammals, black, and live in forests}, expressed in the form of subject-relation-object triples, e.g., \texttt{(gorilla, AtLocation, forest)}. This allows QA systems, for instance, to answer ``\textit{Where do gorillas live?}''. 
On the other hand, CSKBs hardly capture any negative statements
such as ``\textit{gorillas are not territorial}'' or ``\textit{gorillas are not carnivorous}''. 
By the Open-world Assumption (OWA) underlying most KBs, one cannot assume that an absent statement is invalid~\cite{FHPPW2006}; instead, its truth is simply \textit{unknown}. While KB completion~\cite{WPKD2020,DPWQ+2019,Malaviya_Bhagavatula_Bosselut_Choi_2020} is an active research area, creating an ideal KB that fully represents real-world knowledge is elusive, especially for the case of commonsense assertions~\cite{DBLP:journals/ftdb/WeikumDRS21}. Therefore, QA over KBs cannot answer ``\textit{Are gorillas territorial?}''. 
However, such \textit{uncommon knowledge} has value for robust AI applications, asserting that
gorillas are \textit{not} territorial, unlike other apes (and monkeys)
like chimpanzees or gibbons.

\noindent
\textbf{State of the art and its limitations. } The focus in constructing CSKBs has been on positive statements; only very few projects capture a small fraction of negative statements. In ConceptNet~\cite{conceptnet}, a crowdsourced KB, 6 negative relations are represented, namely NotIsA, NotCapableOf, NotDesires, NotHasA, NotHasProperty, and NotMadeOf.
Nonetheless, in its latest version, the portion of negative statements is less than 2\%.
%
Moreover, 
many statements are all but informative, such as \texttt{(envelope, NotHasProperty, alive)}.
In the automatically constructed, Web-based CSKB Quasimodo~\cite{quasimodo}, 350k negated statements represent about 10\% of all statements, 
but these are dominated by uninformative knowledge, e.g., \texttt{$\neg$(elephant, can, quit smoking)}.
A recent method that targets the problem of discovering relevant commonsense negations is NegatER~\cite{safavi-etal-2021-negater,negaterworkshop}. Given a CSKB and a pre-trained language model (LM), e.g., BERT~\cite{BERT}, in order to strengthen the LM's ability to classify true and false statements, the LM is first fine-tuned using the CSKB statements. In a second step, plausible negation candidates are generated using dense k-nearest-neighbors retrieval, by either replacing the subject or the object with a neighboring phrase.
In a final step, the set of plausible candidates are ranked, using the fine-tuned LM, by descending order of \textit{negativeness} (i.e., higher scores are more likely to be negative).
Even though NegatER compiles lists of thematically-relevant negations, it suffers from several limitations: 
(i) The taxonomic hierarchy between concept phrases is not considered. For instance, from the positive statement \texttt{(horse, IsA, expensive pet)}, a semantically sensible corruption of the subject is \texttt{hamster}, but \texttt{horse riding} or \texttt{horserider} are not. Even though they are closer in embedding space, they describe concepts of completely different types (activity, artifact) that cannot be pets.
(ii) The method relies on the input CSKB having 
well-defined relations (e.g., \texttt{CapableOf}). This causes issues when triples are merely short phrases with no canonicalized relations (e.g., as in the Quasimodo CSKB); 
(iii) The ranking based on the LM's negativeness prediction is not interpretable, and follows no clear trend.


\noindent
\textbf{Approach and contributions. } This paper presents the \uncommon{} method for identifying \textit{informative negations} about concepts in CSKBs. For a target concept like \texttt{gorilla}, we first compute a set of comparable concepts (e.g., \texttt{lion}, \texttt{zebra}), by employing both structured taxonomies and latent similarity. Among these concepts, we postulate a Local Closed-world Assumption (LCWA) \cite{galarraga2017predicting}, and consider their positive statements that do not hold for the target concept as candidate negations (e.g, has tail, is territorial). 
To eliminate false positives,
candidates are scrutinized against related statements in the input KB using sentence embeddings, and against a pre-trained LM acting as an external source of latent knowledge. In a final step, we quantify the informativeness of negative statements by statistical scores, and generate top-ranked negations with provenances showing why certain negations are interesting. For instance, \texttt{$\neg$(gorilla, has, tail)}, unlike other \textit{land mammals}, e.g., \texttt{lion} and \texttt{zebra}.

The salient contributions of this work are:
\begin{enumerate}
    \item We present a method for identifying {\em informative} negations about everyday concepts in large-scale CSKBs grounded in taxonomic hierarchies between concepts.
    \item We showcase the ability of our method to produce
    \textit{interpretable} negations via human-readable phrases.
    \item In intrinsic evaluations, our method achieves up to 18\% improvement in informativeness and 17\% in recall, 
    compared to the prior state-of-the-art.
    \item In three extrinsic evaluations, (i) trivia summaries, (ii) KB completion, and (iii) multiple-choice QA, our method 
    shows substantial improvements
    in informativeness.
    \item We release the first large dataset of informative commonsense negations, containing over 6 Million negative statements about 8,000 concepts.\footnote{\url{https://www.mpi-inf.mpg.de/departments/databases-and-information-systems/research/yago-naga/commonsense/uncommonsense}}\\

\end{enumerate}

\section{Problem and Design Space}
\label{sec:preli}
A commonsense KB consists of a finite set of statements in the form  \texttt{(s, r, t)}, where \texttt{s} is a subject (or concept), \texttt{r} is a relation, and \texttt{t} is a tail phrase. Following previous work~\cite{chalierjoint}, we do not distinguish between \texttt{r} and \texttt{t}, because for textual CSK expressions, these distinctions are often ad-hoc and a crisp definition of relations is difficult. Hence, for the remainder of the paper, we generalize the above form to \texttt{(s, f)}, where \texttt{s} is the subject and \texttt{f} is a short \underline{ph}rase combining \texttt{r} and \texttt{t}. 
\begin{definition}
A commonsense negative statement \texttt{$\neg$(s, f)}, is a statement \texttt{(s, f)} that is \textit{not} true.
\end{definition}
For example, ``\textit{elephants are not carnivorous}'' is expressed as \texttt{$\neg$(elephant, is carnivore)}. One naive approach to produce such negations is to assume the CWA (closed-world assumption) over the KB and consider all non-existing statements as negatives. On top of not being materializable, this approach faces the following challenges. \textbf{C1: Avoid false negatives.} In order to assert a negation, it is not sufficient to check if a candidate negation is not positive. KBs in general operate under the OWA (open-world assumption), which means that absent information is merely \textit{unknown}, and not necessarily false. For example, in Ascent, the absence of statement \texttt{(elephant, has eye)} is clearly due to missing information. 
\textbf{C2: Generate judgeable negations.} Whether constructed using human crowdsourcing~\cite{conceptnet} or information extraction techniques~\cite{ascentpp,quasimodo}, KBs mainly reflect the ``wisdom'' of the crowd about everyday concepts. This causes the augmentation of many subjective or otherwise uninformative statements, such as \texttt{(cat, is important)} and \texttt{(football, is boring)}. A generated negation must be easily interpreted by a human annotator as true or false. Therefore it is important to clean the candidate space prior to materializing negations. 
\textbf{C3: Generate informative negations}. Finally, the explicit materialization of all possible negations is not necessary for most standard AI applications (e.g., user might confuse \texttt{tabbouleh} as something that requires an oven but not a \texttt{printer}). In other words, it is better to avoid nonsensical negative statements such as \texttt{$\neg$(printer, is baked in oven)}.\\

\noindent
\textbf{Research problem. } Given a target concept \texttt{s} in a CSKB, generate a ranked list of \textit{truly negative} and \textit{informative} statements.

\section{The UnCommonSense Framework}
\label{sec:methodology}
\begin{figure*}
\includegraphics[width=1.0\linewidth]{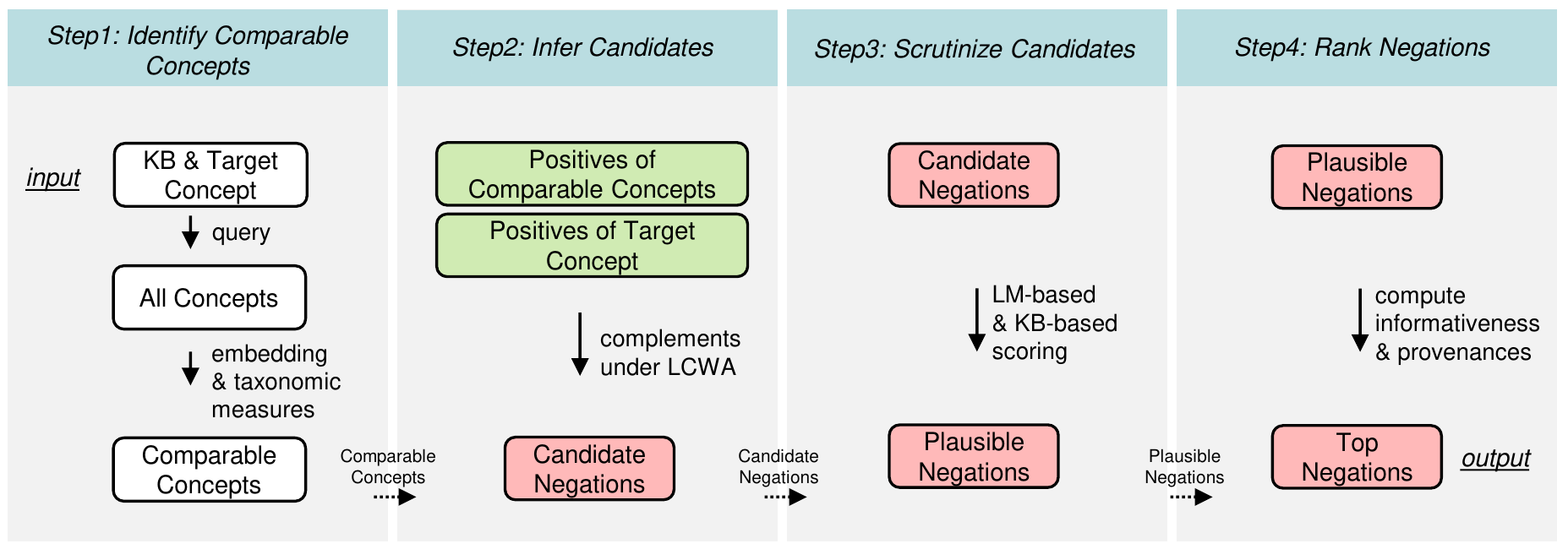}
\caption{Architecture of \uncommon{}.} \label{method}
\end{figure*}

We present \uncommon{}, a method for automatically identifying informative negative knowledge about everyday concepts. \uncommon{} first retrieves comparable concepts for a target concept \texttt{s} by exploiting embeddings and taxonomic relations between these concepts. Over the positive knowledge about these comparable concepts, a \textit{local} closed-world assumption (LCWA) \cite{galarraga2017predicting} is made. These relevant positives are then considered as potential informative  negations for \texttt{s}. Consequently, these candidates might contain many false negatives and nonfactual statements. This is followed by an inspection step, where we use KB-based and LM-based checks to measure the plausibility of candidates. Finally, to measure informativeness, the remaining candidates are scored using relative frequency. An overview is shown in Figure~\ref{method}.

\subsection{Identifying Comparable Concepts}
\label{sec:context}
To increase the thematic relevance of candidate negations, we define the parts of the KB where the CWA is helpful to assume \cite{galarraga2017predicting}, i.e., the LCWA. For instance, if the target concept is an animal, negations should mostly reflect animal-related statements such as ``\textit{not carnivorous}'' or ``\textit{not nocturnal}'', instead of ``\textit{not beverage}'' or ``\textit{cannot store data}''. Therefore, we need to collect \textit{comparable} concepts \cite{recoin}. One way for collecting related concepts is using pre-computed embeddings. For instance, \texttt{elephant} is related to both \texttt{tiger} and \texttt{lion}, due to their proximity in the vector space~\cite{KGembsurvey}. The problem with relying solely on this similarity function is that it does not take into consideration the taxonomic hierarchy of the concepts. For example, \texttt{trunk}, \texttt{circus}, and \texttt{jungle} are also highly related to \texttt{elephant}. Instead, one can consider using large collections of taxonomic relations and collect comparable concepts only if they are listed as co-hyponyms (e.g., \texttt{lion} and \texttt{elephant} are, \texttt{trunk} and \texttt{elephant} are not). Although this option ensures that related concepts are taxonomic sibling, the group of siblings is unordered. For instance, even though \texttt{lion} and \texttt{spider} are both acceptable taxonomic siblings (under \textit{animal}), one is clearly more related to \texttt{elephant} than the other. Moreover, large-scale taxonomies are noisy. For instance, using WebIsALOD~\cite{webisalod}, \texttt{elephant} and \texttt{robot} are co-hyponyms under the class \textit{toy}. We overcome these limitations by combining both techniques and compute \textit{comparable} concepts that are both \textit{semantically and taxonomically highly related}. Given concept \texttt{s}:
\begin{enumerate}
    \item Using latent representations~\cite{wikipedia2vec}, we compute the cosine similarity score between embeddings of \texttt{s} and every other concept in the KB, and rank them by descending order of similarity.
    \item Using hypernymy relations~\cite{wang-etal-2017-short}, we retain siblings that are co-hyponyms of \texttt{s}. In particular, for every concept, we collect the top-5 hypernyms (ranked by confidence score\footnote{Using WebIsALOD's SPARQL endpoint: \url{https://webisadb.webdatacommons.org/}}). For instance, \texttt{elephant} has 843 hypernyms. Top ones include \textit{larger animal}, \textit{land animal}, and \textit{mammal}, and bottom ones include \textit{work of art}, \textit{african}, and \textit{symbol of power}. We retain KB concepts as comparable to our target concept if they pass the following taxonomic checks: (i) There exists a common hypernym with the target concept (e.g., both \texttt{elephant} and \texttt{tiger} share \textit{mammal}), and (ii) There does \textit{not} exist an \texttt{IsA} relation with the target concept, e.g., \texttt{african elephant, IsA, elephant}, hence \texttt{african elephant} is not a valid sibling.
\end{enumerate}
The ideal number of comparable concepts to consider for every target concept is a hyperparamater $\gamma$, which we tune in Section~\ref{sec:ablation}. For the remainder of the paper, we use the terms \textit{comparable concepts} and \textit{siblings} interchangeably. 

\example{Given \texttt{s} = \texttt{elephant} from KB = \textit{Ascent}, and $\gamma$ = 3, the concepts with the highest cosine similarity are computed using Wikipedia2Vec~\cite{wikipedia2vec}. The ranked candidate concepts include \texttt{tiger, lion, trunk, horse, $\dots$}. Here, \texttt{trunk} is an obvious intruder as it is not share a hypernym with \texttt{s}. This is determined using WebIsALOD~\cite{webisalod,madoc47550}, an Is-A database, containing 400m hypernymy relations, mined, using over 50 Hearst-style patterns, from a huge web crawl. We end up with the closest 3 siblings: \texttt{tiger}, \texttt{lion}, and \texttt{horse}.
This initial step is meant to address challenge \textbf{C3}, which we further demonstrate in Section~\ref{sec:ablation} (Table~\ref{tab:ablation}).}

 \subsection{Candidate Negation Inference}
To produce a set of candidate negations, we query from the KB the set of positives about \texttt{s} as well as positives about its siblings. We subtract both sets to produce an initial set of candidates $N$:
\begin{equation}
\label{eqn:infer}
N = B \ \backslash \ A
\end{equation}
where $B$ is the set of phrases describing sibling concepts (i.e., each phrase holds for at least one sibling), and $A$ is the set of phrases that hold for the target concept. So, $N$ contains phrases that are $\in$ $B$ but $\notin$ $A$. 

\example{\texttt{elephant}'s statements (i.e., $A$) are: \texttt{(is largest land animal)} and \texttt{(has tongue)}. Positives of the siblings (i.e., $B$) are \texttt{(is amazing)},  \texttt{(can jump)}, \texttt{(has tongue)}, \texttt{(has hoof)}, \texttt{(eat grass)}, \texttt{(can leap)}, and \texttt{(is big animal)}. The negation set $N$ is then all the phrases in the siblings' set except for \texttt{has tongue}, which is a straightforward contradiction with positives about \texttt{elephant}.}
 
 \subsection{Scrutinizing Candidates}

\noindent
\textbf{Plausibility checks. } To remove candidates that might be inaccurate due the KB's incompleteness, and address \textbf{C1}, we measure the plausibility of our candidate negations in two steps:
\begin{enumerate}
    \item \textit{KB-based scoring}: Unlike encyclopedic KB (e.g., Wikidata~\cite{WD}), statements in CSKB are semi-structured. Therefore, it is possible that the same piece of information is expressed in various ways. For example, \textit{lay eggs}, \textit{deposit eggs}, and \textit{lie their eggs} are phrases that hold for different insects in Ascent. Our simple set difference will miss such contradictions. To overcome this issue, we exploit sentence-embeddings~\cite{reimers-2019-sentence-bert} to capture semantically-close statements in the KB, namely semantically-close information between the concept's and siblings' positives. We filter out candidates that are highly similar to information we already know about the target concept. 
    \item \textit{LM-based scoring}: In open-world KBs, it is not sufficient to perform a plausibility check against the knowledge in the KB, as valuable statements might be simply \textit{missing}. We propose consulting an external source for further investigation of candidates. In particular, we probe LMs in a zero-shot manner for factual knowledge~\cite{DBLP:journals/corr/abs-2110-04888}, by masking the target concept and concatenating the candidate phrase. We then look for a match between predicted tokens and the unmasked concept. We only mask the target concept since it is the most decisive part of a statement. 
\end{enumerate}

\example{Using Sentence-BERT (or SBERT~\cite{reimers-2019-sentence-bert}), we measure the similarity (sim) of the candidate and positive phrases:
\begin{itemize}
\item[] sim(``can jump'', ``is largest land animal'') = 0.05
\item[] sim(``can jump'', ``has hoof'') = 0.20
\item[] ...
\item[] sim(``is big animal'', ``is largest land animal'') = \textbf{0.78}
\end{itemize}
The candidates with similarity greater than or equal a certain threshold $\lambda$ (in this example 0.7) are considered \textit{false negatives}. In this case, we drop the candidate \texttt{$\neg$(elephant, is big animal)}. 

Next, using BERT~\cite{BERT}, we construct a probe with masked target concept concatenated with the candidate phrase and look for \texttt{s} in the first $\tau$ predictions (in this example 100) as follows.
\begin{itemize}
\item[] [MASK] has hoof. (\textit{no ``elephant'' in top-100})
\item[] [MASK] can jump. (\textit{no ``elephant'' in top-100})
\item[] ...
\item[] [MASK] eat grass. (\textbf{``elephants'' at position 76})
\end{itemize}
In this case \texttt{$\neg$(elephant, eat grass)} is dropped from the candidate set.}

\noindent
\textbf{Quality checks.} To avoid vague or opinionated negations such as \texttt{$\neg$(classroom, is bigger)} or \texttt{$\neg$(basketball, is important)}, and address \textbf{C2}, we identify frequent statements that are highly uninformative. Inspired by the notion of term-weighting in IR~\cite{manning} (in our case, phrase-weighting), we value phrases of \textit{medium-frequency}, namely ones that are neither too generic nor too rare. While we ensure that rare statements are lower ranked via the pipeline's final step, we tackle too generic statements follows: A statement is \textit{generic} if it holds for $\geq$ $\beta$ of the concepts in the KB.
 
 \example{With $\beta$ = 0.05, \texttt{$\neg$(elephant, is amazing)} is dropped from the candidate set as it holds for 16\% ($\geq$ 5\%) of \textit{all} the concepts in Ascent.} 
 
 Hyperparameters $\lambda$, $\tau$, and $\beta$ are tuned in Section~\ref{sec:ablation}.

\subsection{Quantifying Informativeness}

 The output of the previous step is a potentially large set of \textit{truly negative} statements. In fact, beyond our toy example, starting with 30 siblings for \texttt{elephant}, \uncommon{} produces 1352 initial candidates. Hence, ranking is crucial. We quantify the \textit{importance} of a certain candidate negation by how \textit{uncommon} it is among its siblings. The notion of informativeness is expressed through unique behavior, characteristic, and so on, of a certain concept, given what is known about its siblings. More formally, given a candidate phrase \texttt{f} about target concept \texttt{s} and its siblings $\{x_{1}, x_{2},.., x_{\gamma}\}$, we measure \texttt{f}'s informativeness using \textit{strict sibling frequency}.
 
\begin{equation}
\label{eqn:strict}
\textit{strict}(\text{\texttt{s}}, \text{\texttt{f}}, \{x_{1}, x_{2},.., x_{\gamma}\}) = \frac{|\{x_{i} | (x_{i}, \text{\texttt{f}}) \in \text{KB}\}|}{\gamma}
\end{equation}

\example{To score candidates, we compute: \textit{strict}(\texttt{elephant}, \texttt{has hoof}, $\{$\texttt{tiger, lion, horse}$\}$) = $|\{$\texttt{horse}$\}|$/3 = 0.33, \textit{strict}(\texttt{ele\-phant}, \texttt{can jump}, $\{$\texttt{tiger, lion, horse}$\}$) = $|\{$\texttt{tiger, horse}$\}|$/3 = 0.67, and \textit{strict}(\texttt{elephant},  \texttt{can leap}, $\{$\texttt{tiger, lion, horse}$\}$) = $|\{$\texttt{lion}$\}|$/3 = 0.33. Therefore it is more noteworthy that elephants cannot jump, unlike \textit{all} their siblings.}

\noindent
\textbf{Relaxed scoring. } The \textit{strict} informativeness scoring only handles the cases where candidate negations are expressed using the same exact phrasing. It cannot, however, capture cases where highly similar candidates are stated using different wording. For instance, the candidate set might contain both \texttt{$\neg$(elephant, can jump)} and \texttt{$\neg$(elephant, can leap)}. To remedy this, we make use of sentence embeddings~\cite{reimers-2019-sentence-bert} in order to capture this similarity and boost the scores of candidates. We measure \texttt{f}'s informativeness using \textit{relaxed sibling frequency} as follows.

\begin{equation}
\label{eqn:relaxed}
\begin{aligned} 
\textit{relaxed}(\text{s}, \text{\texttt{f}}, \{(x_{1},[\text{\texttt{f$^{\prime 1}_1$}},\dots]),\dots, (x_{\gamma},[\text{\texttt{f$^{\prime 1}_\gamma$}},\dots])\}) = \\ 
\frac{|\{ x_{i} | (x_{i}, \text{\texttt{f$^{\prime j}_i$}}) \in \text{KB} \land (\text{\texttt{f}} = \text{\texttt{f$^{\prime j}_i$}} \vee (sim(\text{\texttt{f}}, \text{\texttt{f$^{\prime j}_i$}}) \geq \lambda))\}|}{\gamma}
\end{aligned} 
\end{equation}

where \text{\texttt{f$^{\prime j}_i$}} is a phrase that holds for sibling $x_{i}$ and $sim(\text{\texttt{f}}, \text{\texttt{f$^{\prime j}_i$}})$ is the semantic similarity between candidate \texttt{f} and candidate-rephrase \text{\texttt{f$^{\prime j}_i$}}.

\example{Candidates \texttt{(can jump)} and \texttt{(can leap)} are semantically similar, hence they are combined under the relaxed scoring. In particular, we compute: \textit{relaxed}(\texttt{elephant}, \texttt{can jump}, $\{$(\texttt{horse}, [\text{\texttt{can jump}}]), (\texttt{lion}, [\text{\texttt{can leap}}]), (\texttt{tiger}, [\text{\texttt{can jump}}])\}) = $|\{$\texttt{tiger, lion, horse}$\}|$/3 = 1.0. 
}

\noindent
\textbf{Provenance generation. } Unlike in previous work \cite{safavi-etal-2021-negater}, negations generated by \uncommon{} come naturally with an explanation via the relationship between the siblings and the target concept. We call these explanations \textbf{\textit{negation provenances}}. We generate these human-readable phrases by measuring the in-group frequency of shared hypernyms. In particular, we compute a score for each hypernym $h$ that holds for \texttt{s} within the set of siblings sharing phrase \texttt{f}.

\begin{equation}
\label{eqn:prov}
score(h, \text{\texttt{s}}, \text{\texttt{f}}, \{x_{1}, x_{2},.., x_{n}\}) = \frac{|\{x_{i} | (x_{i}, \text{isA}, h) \in \text{TX}\}|}{n}
\end{equation}
where TX is the taxonomic relations database, e.g., WebIsALOD~\cite{webisalod}, ($x_{i}$, isA, $h$) $\in$ TX indicates that hypernym $h$ holds for sibling $x_{i}$ in TX, and $n$ is the total number of siblings candidate-phrase \texttt{f} holds for.

\example{Assume \texttt{elephant} has the hypernyms \textit{wild mammal} and \textit{herbivorous animal}. To build the provenance for top negation \texttt{$\neg$(elephant, can jump)} which holds for all siblings (by relaxed scoring) we compute: score(\textit{wild mammal}, \texttt{elephant}, \texttt{can jump}, $\{$\texttt{tiger, horse, lion}$\}$) =  $|\{$\texttt{tiger, lion}$\}|$/3 = 0.67, and score(\textit{herbivorous animal}, \texttt{elephant}, \texttt{can jump}, $\{$\texttt{tiger, horse, lion}$\}$) =  $|\{$\texttt{horse}$\}|$/3 = 0.33. The provenance-extended negation then reads: \texttt{$\neg$(elephant, can jump)} unlike other \textit{wild mammals}, e.g., tiger, lion, and unlike other \textit{herbivorous animals}, e.g., horse.

\medskip 

\noindent To avoid potential multiple appearances of siblings in one provenance, i.e., one sibling belonging to several subgroups, we compute $h$ with the highest score iteratively such that at every iteration we drop already seen siblings.}

\section{Experiments}
\label{sec:experiments}
The evaluation of \uncommon{} is centered around answering the 3 challenges introduced in Section~\ref{sec:preli}. Hence, we conduct:
\begin{enumerate}
    \item An intrinsic evaluation to demonstrate the ability of our method to produce plausible (\textbf{C1, C2}) and informative (\textbf{C3}) commonsense negative knowledge against baseline and state-of-the-art methods. We demonstrate the ability of our method to extend negations with valuable information (\textbf{C3}) by an evaluation of provenances.
    \item Three extrinsic evaluations:
    \begin{enumerate} 
        \item A \textit{negative-trivia} use-case where we evaluate the quality of summaries about concepts (\textbf{C2, C3}).
        \item A \textit{KB completion} use-case where we provide challenging negative examples for LM-based triple classifier (\textbf{C1, C3}).
        \item A \textit{multiple-choice QA} use-case, where we utilize our model as an eliminator to exclude improbable options (\textbf{C1}).
    \end{enumerate}
    
\end{enumerate}

\subsection{Setup}
\textbf{Data Source: Ascent CSKB.}
We use Ascent++~\cite{ascentpp} as our input CSKB (in the following just called Ascent). This choice is motivated by the fact that computing negations benefits from richer input sets (i.e, high statement-recall per concept). In comparison, in ConceptNet, the most prominent CSKB, has 23 statements per concept on average. Ascent, on the other hand, has 256. Moreover, Ascent contains 2m assertions for 23k subjects. We restrict our evaluation to the 8k \textit{primary} subjects and disregard \textit{aspects} and \textit{subgroups}. 

\noindent
\textbf{Baselines and implementation.} We compare our method to the following baselines.
\begin{itemize}
    \item \textbf{CWA}: In this baseline, the KB is simply assumed to be complete. For a given concept, any phrase not asserted gives an immediate negation. 
    \item \textbf{Quasimodo\textsuperscript{neg}}: We download the latest version of Quasimodo~\cite{quasimodo} and retrieve all the statements with \textit{negative polarity} (a total of 350k negations, e.g., \texttt{$\neg$(baby, has hair)}).
    \item \textbf{GPT-3\textsuperscript{neg}}: We prompt GPT-3~\cite{gpt3} daVinci model using predefined prompts with negative keywords. Based on Ascent's relations, we define 10 most frequent relations and map to 8 manually-crafted meta patterns ``<\texttt{s}> <Negated\_NL\_relation> ...''. <Negated\_NL\_relation> stands for negated natural language relations we created by rephrasing Ascent's canonicalized relations, namely ``MadeOf'' to ``is not made of'', ``CapableOf'' to ``cannot'', ``IsA, HasProperty, ReceivesAction'' to ``is not'', ``HasA'' to ``does not have'', ``AtLocation'' to ``is not found in'', ``Causes'' to ``does not cause'', ``HasSubevent'' to ``does not lead to'', and ``HasPrerequisite'' to ``does not need''. A sample prompt is ``butterfly is not \underline{a bird}''. We restrict predictions to a maximum of 6 tokens. We produce 24.4k negations about 200 concepts.
    \item \textbf{NegatER}-$\pmb{\theta}_{r}$~\cite{safavi-etal-2021-negater}: This work presents an unsupervised met\-hod that ranks out-of-KB potential negatives using a fine-tuned LM. We use the released code\footnote{\url{https://github.com/tsafavi/NegatER}} to fine-tune BERT on the full Ascent. Similar to the original implementation on ConceptNet, we divide the Ascent dataset into 1.6m/41k/41k rows for training/validation/test, with a total of 715k entity phrases. The evaluation sets are constructed in the same manner, i.e., in terms of balance and negative sampling. We use the given best configuration file and run the fine-tuning step for 3 epochs (6 hours each), using an NVIDIA Quadro RTX 8000 GPU with 48GB of RAM. On the test set, we obtain precision=0.96, and recall=accuracy=0.97. We run the negation generator first in the ranking version NegatER-${\theta}_{r}$, which relies on decision thresholds.
    \item \textbf{NegatER}-${\nabla}$~\cite{safavi-etal-2021-negater}: We also run the above negation generator in the ${\nabla}$ setting, which relies on quantifying ``surprisal'' using  LM’s gradients. Using both variants of the method, we produce more than 16m scored negations. Note that while we use \textit{canonicalized} Ascent to run NegatER, e.g., \texttt{(elephant, CapableOf, jump)}, for consistency of examples across the methods, we show the \textit{open} version of the triple, e.g., \texttt{(elephant, can jump)}.
\end{itemize}
All the extracted/generated negations of the three external methods are released for future comparisons\footnote{\url{https://www.mpi-inf.mpg.de/departments/databases-and-information-systems/research/yago-naga/commonsense/uncommonsense}}.

\medskip

\noindent
\textbf{\uncommon{} Variants. } We consider three variants.
\begin{itemize}
    \item \textbf{\uncommon{}\textsuperscript{B}}: This baseline variant computes comparable concepts as described in Section~\ref{sec:context}, but suspends the scrutinizing and ranking steps.
    \item \textbf{\uncommon{}\textsuperscript{S}}: The complete method, with informativeness computed using strict ranking (i.e., Equation~\ref{eqn:strict}).
    \item \textbf{\uncommon{}\textsuperscript{R}}: The complete method, with informativeness computed using relaxed ranking (i.e., Equation~\ref{eqn:relaxed}).
\end{itemize}
    For all variants, $\gamma$ is set to 30, $\tau$ to 50, $\lambda$ to 0.7, and $\beta$ to 0.05. These hyperparameters are chosen based on a tuning task in Section~\ref{sec:ablation}.
    Moreover, we collect taxonomic siblings from WebIsALOD~\cite{webisalod} and order them using Wikipedia2Vec~\cite{wikipedia2vec}. We use SBERT~\cite{reimers-2019-sentence-bert} for sentence similarity checks and use BERT~\cite{BERT} for LM-based checks.

\subsection{Intrinsic Evaluation}
\label{sec:intrinsic}

\noindent
\textbf{Human plausibility and informativeness evaluation. } We conduct a human evaluation\footnote{\url{https://www.mturk.com/}} to determine the quality of each method in generating plausible and salient negations. We randomly sample 200 concepts and produce for each top-2 negations. We then acquire 3 annotations for each negation via crowdsourcing. The total number of annotated negations is 200 concepts $\times$ 2 negations $\times$ 8 methods $\times$ 3 annotations = 9.6k rows. We ask every annotator to answer two questions, given a statement about a concept: 1) Is this statement truly negative?, 2) Is this statement interesting and/or useful in your opinion? Since question 1) is a factual question, we only allow ``yes'' and ``no'', which we map to 1 and 0 respectively. The Fleiss' kappa~\cite{fleiss} inter-annotator agreement is 0.46, i.e. moderate agreement. We interpret this slightly underwhelming agreement on this relatively easy task by the large number of \textit{opinionated statements} produced especially using the baseline methods, e.g., \texttt{$\neg$(football, is boring)},  \texttt{$\neg$(muffin, is delicious)}. For question 2), an annotator chooses between ``interesting'', ``slightly interesting'', and ``not interesting'', which we map to 1, 0.5, and 0 respectively. The agreement on this arguably vague task is fair, with Fleiss' kappa inter-annotator agreement 0.30. Numerical results on informativeness and plausibility are shown in Table~\ref{tab:intrinsic} and qualitative examples in Table~\ref{tab:intrinsicexamples}. The \textit{false negatives} column reflects the ratio of result-negations that are in fact positive (i.e., not plausible). Obviously, the CWA baseline dominates with only 0.07\% false inferences, as the majority of the produced negations are accurate but \textit{nonsensical} (e.g., in Table~\ref{tab:intrinsicexamples}, ``\textit{rabbits are not related to bribery}''). On the notion of \textit{informativeness}, the leading method is \uncommon{} in its both ranking variants, outperforming the second best external method, Quasimodo\textsuperscript{neg}, by 18\%, with a slight advantage of the strict ranking variant over the relaxed. We note that in our computation of informativeness, we only consider the negations that have been marked by the majority as truly negative. In this case, only 39\% of the negations proposed by Quasimodo\textsuperscript{neg} are plausible, and even less for GPT-3\textsuperscript{neg} with 37\%, as opposed to 75\% for \uncommon{} and 74\% for NegatER. We observe for the baseline variant \uncommon{}\textsuperscript{B} that negations are mostly thematic (due to inferences based on \textit{comparable} concepts), however not frequent enough (due to absence of ranking), e.g., \texttt{$\neg$(gorilla, caught in net)} and in some cases false (due to absence of candidate-scrutiny), e.g., \texttt{$\neg$(rabbit, can feed on seed)}.
In Table~\ref{tab:intrinsicexamples}, \uncommon{} shows the most interesting results. For example, it is worth noting that unlike many other small mammals, ``\textit{rabbits do not eat insects}''. To give more insights into different kinds of concepts, we show the informativeness of each method per topic. The results are shown in Table~\ref{tab:intrinsicfiner}. \uncommon{} performs best on topics like animal and food with informativeness scores of 67\% and 55\% respectively. This is expected as both themes contain the most factual statements, and are fairly easy to judge e.g., \texttt{$\neg$(banana, is bitter)} and \texttt{$\neg$(horse, eat fruit)}. On the other hand, it is more challenging to judge social negations, e.g., \texttt{$\neg$(niece, is pregnant)} and \texttt{$\neg$(alcoholic, has friend)}. 

\noindent
\textbf{Automated recall evaluation. } To measure recall, we collect top-200 negations, per target concept, produced by each method. Moreover, we need a ground-truth dataset with negative statements about KB concepts. We create \textbf{ConceptNet-neg} by retrieving all the statements from ConceptNet~\cite{speer-havasi-2012-representing} v5.5 that have a negative relation. This KB allows 6 negative relations such as \texttt{NotCapableOf} and \texttt{NotDesires}. The dataset contains 14.1k negations. Samples include \texttt{(butterfly,	NotDesires,	to sting like a bee)} and \texttt{(tortoise, NotIsA, a turtle)}. We remove the negative keywords from relations (i.e., the prefix \texttt{NOT}). We then compute two modes of recall: In the \textit{strict} mode, we consider a generated negation by a given method to be valid if it matches the \textit{exact phrasing} of a negation in the ground-truth. In the \textit{relaxed} mode, we use embedding similarity~\cite{reimers-2019-sentence-bert} to assess whether a generated-negation and a ground-truth-negation are of similar meaning. The recall results are shown in Figure~\ref{fig:recall}. \uncommon{} outperforms all methods.  The strict mode is tougher since the slightest difference between the ground-truth and method-generated negations is considered a mismatch, e.g., \texttt{$\neg$(air conditioner, quiet)} and \texttt{$\neg$(air conditioner, quieter)}. Relaxing the matching rule to sentence similarity~\cite{reimers-2019-sentence-bert} allows for more forgiving comparisons. Our method reaches 26.1\% in relaxed@10 (relaxed recall at top-10 negations), followed by NegatER-${\nabla}$ with 9.6\%, Quasimodo\textsuperscript{neg} with 6.3\%, and finally GPT-3\textsuperscript{neg} with 4.4\%. An example of a relaxed match is the pair of statements \texttt{$\neg$(bicycle, has motor)} (in ground-truth) and \texttt{$\neg$(bicycle, has engine)} (generated by \uncommon{}). 
\begin{table}
\caption{Plausibility and informativeness evaluation.}\label{tab:intrinsic}
\resizebox{\columnwidth}{!}{\begin{tabular}{|l|c|c|}
\hline
\textbf{\cellcolor{gray!15}Method} &  \textbf{\cellcolor{gray!15}False Negatives} & \textbf{\cellcolor{gray!15}Informativeness} \\
\hline
CWA & \underline{0.07}  & 0.07   \\
Quasimodo\textsuperscript{neg}  & 0.61   & 0.32 \\
GPT-3\textsuperscript{neg} & 0.63 & 0.30 \\
NegatER-${\theta}_{r}$  & 0.27  & 0.28 \\
NegatER-${\nabla}$  & 0.26  & 0.29 \\
\hdashline
\uncommon{}\textsuperscript{B} & 0.29  & 0.30  \\
\uncommon{}\textsuperscript{S}  & 0.25  & \underline{0.50}  \\
\uncommon{}\textsuperscript{R}  & 0.27  & 0.47 \\
\hline
\end{tabular}}
\end{table}

\begin{table*}
\caption{Intrinsic evaluation, sample results.}\label{tab:intrinsicexamples}
\begin{tabular}{|l|l|c|}
\hline
\textbf{\cellcolor{gray!15}Method}  & \textbf{\cellcolor{gray!15}Top negations} & \textbf{\cellcolor{gray!15}Truly Negative?} \\
\hline
\multirow{4}{*}{CWA} & $\neg$(acne, can give an understanding of truth) & \textbf{\cmark}\\
& $\neg$(elephant, can provide clinician) & \textbf{\cmark}\\
& $\neg$(yawning, has fluid)& \textbf{\cmark}\\
& $\neg$(vinegar, can comprise about 55\% nickel)& \textbf{\cmark}\\
& $\neg$(rabbit, related to bribery)& \textbf{\cmark}\\
\hline
\multirow{4}{*}{Quasimodo\textsuperscript{neg}} & $\neg$(acne, is natural)& \textbf{\xmark}\\
& $\neg$(elephant, quit smoking)& \textbf{\cmark}\\
& $\neg$(yawning, can end)&  \textbf{\xmark}\\
& $\neg$(vinegar, is vegan)& \textbf{\xmark}\\
& $\neg$(rabbit, is rodent)& \textbf{\cmark}\\
\hline
\multirow{4}{*}{GPT-3\textsuperscript{neg}} & $\neg$(acne, can be cured)& \textbf{\cmark}\\
& $\neg$(elephant, found in the dictionary)& \textbf{\xmark}\\
& $\neg$(yawning, can be controlled)&  \textbf{\xmark}\\
& $\neg$(vinegar, need to be refrigerated) & \textbf{\cmark}  \\
& $\neg$(rabbit, found in the wild)& \textbf{\xmark}\\
\hline
\multirow{4}{*}{NegatER} & $\neg$(acne, become unresponsive)& \textbf{?}\\
& $\neg$(elephant, interested)& \textbf{?} \\
& $\neg$(yawning, attenuated by atropine)& \textbf{\cmark}\\
& $\neg$(vinegar, stocked with herb) & \textbf{\cmark}\\
& $\neg$(rabbit, is the most important animal)&  \textbf{?} \\
\hline
\multirow{4}{*}{\uncommon{}} & $\neg$(acne, is fatal)& \textbf{\cmark}\\
& $\neg$(elephant, is carnivore)  & \textbf{\cmark}\\
& $\neg$(yawning, can relax muscles)  & \textbf{\cmark}\\
& $\neg$(vinegar, has iron) & \textbf{\cmark}\\
& $\neg$(rabbit, eat insect) & \textbf{\cmark}\\
\hline
\end{tabular}
\end{table*}

\begin{figure}
\scriptsize
\begin{subfigure}{\columnwidth}
\centering
\begin{tikzpicture}
\begin{axis}[
    ybar,
    ymin=0,
    bar width=0.30cm,
    height=4cm,
    width=.95\columnwidth,
    enlarge x limits=0.5,
    ylabel={recall (\%)},
    symbolic x coords={strict,strict$@$10},
    xtick=data,
    ]
\addplot[draw = black,line width = .2mm,fill = pink,postaction={pattern=north west lines}] 
    coordinates {
        (strict, 0.5)
        (strict$@$10, 0.3)
    }; \label{QUASI-bar}
\addplot[draw = black,line width = .2mm,fill = orange,postaction={pattern=grid}] 
    coordinates {
        (strict, 0.5)
        (strict$@$10, 0.6)
    }; \label{GPT-bar}
\addplot[draw = black,line width = .2mm,fill = green,postaction={pattern=crosshatch}] 
    coordinates {
        (strict, 0.1)
        (strict$@$10, 0.1)
    }; \label{negatertheta-bar}
\addplot[draw = black,line width = .2mm,fill = cyan,postaction={pattern=vertical lines}] 
    coordinates {
        (strict, 1.5)
        (strict$@$10, 1.5)
    }; \label{negaternabla-bar}
\addplot[draw = black,line width = .2mm,fill = yellow,postaction={pattern=crosshatch dots}] 
    coordinates {
        (strict, 5.2)
        (strict$@$10, 5.2)
    }; \label{uncommonbase-bar}
\addplot[draw = black,line width = .2mm,fill = red,postaction={pattern=horizontal lines}] 
    coordinates {
        (strict, 7.6)
        (strict$@$10, 7.6)
    }; \label{uncommonstrict-bar}
\addplot[draw = black,line width = .2mm,fill = lightgray,postaction={pattern=north east lines}] 
    coordinates {
        (strict, 7.8)
        (strict$@$10, 7.9)
    }; \label{uncommonrelaxed-bar}
\end{axis}
\end{tikzpicture}
\quad
\begin{tikzpicture}
\begin{axis}[
    ybar,
    ymin=0,
    bar width=0.30cm,
    height=4cm,
    width=.95\columnwidth,
    enlarge x limits=0.5,
     ylabel={recall (\%)},
    symbolic x coords={relaxed,relaxed$@$10},
    xtick=data,
    ]
\addplot[draw = black,line width = .2mm,fill = pink,postaction={pattern=north west lines}] 
    coordinates {
        (relaxed, 6.3)
        (relaxed$@$10, 6.3)
    };
\addplot[draw = black,line width = .2mm,fill = orange,postaction={pattern=grid}] 
    coordinates {
        (relaxed, 4.2)
        (relaxed$@$10, 4.4)
    };
\addplot[draw = black,line width = .2mm,fill = green,postaction={pattern=crosshatch}] 
    coordinates {
        (relaxed, 5.5)
        (relaxed$@$10, 6.2)
    };
\addplot[draw = black,line width = .2mm,fill = cyan,postaction={pattern=vertical lines}] 
    coordinates {
        (relaxed, 8.6)
        (relaxed$@$10, 9.6)
    };
\addplot[draw = black,line width = .2mm,fill = yellow,postaction={pattern=crosshatch dots}] 
    coordinates {
        (relaxed, 25.9)
        (relaxed$@$10, 25.8)
    };
\addplot[draw = black,line width = .2mm,fill = red,postaction={pattern=horizontal lines}] 
    coordinates {
        (relaxed, 25.5)
        (relaxed$@$10, 26.1)
    };
\addplot[draw = black,line width = .2mm,fill = lightgray,postaction={pattern=north east lines}] 
    coordinates {
        (relaxed, 20.5)
        (relaxed$@$10, 21.3)
    }; 
\end{axis}
\end{tikzpicture}
\label{fig:recalla}
\end{subfigure}

\begin{subfigure}{\columnwidth}
\centering
    \caption*{
        \footnotesize
        \ref{QUASI-bar} Quasimodo\textsuperscript{neg}
        \enspace
        \ref{GPT-bar} GPT-3\textsuperscript{neg}
        \enspace
        \ref{negatertheta-bar} NegatER-${\theta}_{r}$
        \enspace
        \ref{negaternabla-bar} NegatER-${\nabla}$
    }
    \caption*{
        \footnotesize
        \ref{uncommonbase-bar} \uncommon{}\textsuperscript{B}
        \enspace
        \ref{uncommonstrict-bar} \uncommon{}\textsuperscript{S}
        \enspace
        \ref{uncommonrelaxed-bar} \uncommon{}\textsuperscript{R}
    }
\end{subfigure}
\caption{Recall evaluation.}
\label{fig:recall}
\end{figure}
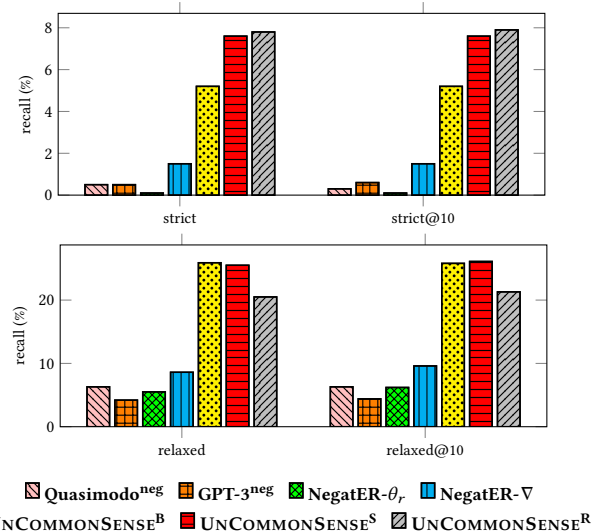
\begin{table*}
\caption{Informativeness per domain.}\label{tab:intrinsicfiner}
\begin{tabular}{|l|c|c|c|c|c|c|}
\hline
\textbf{\cellcolor{gray!15}Method} &  \textbf{\cellcolor{gray!15}Animal} & \textbf{\cellcolor{gray!15}Food} & \textbf{\cellcolor{gray!15}Activity} & \textbf{\cellcolor{gray!15}Human} & \textbf{\cellcolor{gray!15}Object} &  \textbf{\cellcolor{gray!15}Other}\\
\hline
CWA  & 0.06 &0.09   &0.21 & 0.18 & 0.10 & 0.15 \\
Quasimodo\textsuperscript{neg} & 0.41  & 0.44  & 0.24 & \underline{0.39}  & 0.20  & 0.24 \\
GPT-3\textsuperscript{neg}  & 0.14  & 0.46  & 0.44  & 0.17  & 0.22  &0.23 \\
NegatER-${\theta}_{r}$& 0.10 & 0.11 & 0.16 & 0.26 & 0.15 &0.17 \\
NegatER-${\nabla}$& 0.13 & 0.14 & 0.17 & 0.23 & 0.12 &0.18  \\
\hdashline
\uncommon{}\textsuperscript{B}   & 0.29  & 0.37  & 0.32 & 0.24 & 0.24  & 0.27 \\
\uncommon{}\textsuperscript{S}  & \underline{0.67} & 0.52  & \underline{0.49} & \underline{0.39}  & \underline{0.42} & \underline{0.45} \\
\uncommon{}\textsuperscript{R}  & 0.61  & \underline{0.55}  &0.42 & 0.35 & 0.41 &0.42 \\
\hline
\textit{Sample concept} & \textit{lynx} & \textit{waffle} & \textit{basketball} & \textit{niece} & \textit{tripod} & \textit{propaganda}\\
\hline
\end{tabular}
\end{table*}


\begin{table*}
\caption{Examples of provenance-extended negations (\uncommon{}\textsuperscript{V}).}\label{tab:provenance}.
\begin{tabular}{|l|l|}
\hline
\textbf{\cellcolor{gray!15}Target Concept} &  \textbf{\cellcolor{gray!15}Negation}\\
\hline
muffin & $\neg$(is runny) unlike other \textit{breakfast item}, e.g., \textit{syrup,  yogurt}\\
gorilla & $\neg$(is territorial) unlike other \textit{wild animal}, e.g., \textit{tiger, lion, monkey, chimpanzee}\\
vinegar & $\neg$(has iron) unlike other \textit{ingredient}, e.g., \textit{fennel, celery, fenugreek} and \textit{acidic food}, e.g., \textit{tomato}\\
ear & $\neg$(is muscular) unlike other \textit{body part}, e.g., \textit{shoulder, loin, neck}\\
\hline
\end{tabular}
\end{table*}

\begin{table*}
\caption{Example of MCQA through elimination process ({\colorbox{red!30}{eliminated choice}} and \underline{correct choice}).}\label{tab:eliminationexamples}
\begin{tabular}{|l|}
\hline
\multicolumn{1}{|c|}{\cellcolor{gray!15}\textbf{Concept =} hand, \textbf{Query =} What is a hand?}\\
\hline
\textbf{Eliminator = NegatER}\\
\multicolumn{1}{|c|}{{\textbf{A.} foot} {(-)} {\textbf{B.} feet} {(-)}
{\textbf{C.} digestive organ} {(-)}
\textbf{D.} \underline{body part} (-)
\textbf{E.} help (-)}\\
\hline
\textbf{Eliminator = \uncommon{}}\\
\multicolumn{1}{|c|}{{\colorbox{red!30}{\textbf{A.} foot}} { ($\neg$ foot)} {\colorbox{red!30}{\textbf{B.} feet}} { ($\neg$ foot)}
{\colorbox{red!30}{\textbf{C.} digestive organ}} { ($\neg$ digestive system)}
\textbf{D.} \underline{body part} (-)
\textbf{E.} help (-)}\\
\hline
\end{tabular}
\end{table*}

\noindent
\textbf{Evaluation of provenance generation. } To show the effect of extending negation with provenances, we conduct a crowdsourcing experiment to compare \uncommon{} against provenance-extended \uncommon{}. We call the latter \uncommon{}\textsuperscript{V}, as in \textit{verbose}. For 200 concepts, for each variant we produce top-5 negations. The results are then judged by 3 annotators. We ask about the general informativeness of the negations and allow ``interesting'', ``slightly interesting'', and ``not interesting''. \uncommon{}\textsuperscript{V} outperforms \uncommon{} by 32\% in informativeness, with 81\% and 49\% respectively. Examples are shown in Table~\ref{tab:provenance}. The Fleiss' kappa inter-annotator agreement of this task is 0.44, i.e., moderate.

\subsection{Extrinsic Evaluation I: Negative Trivia}
\label{sec:entitysummarization}
Trivia is an umbrella term for interesting knowledge without a specific purpose. We compare methods for negation generation in their ability to generate \textit{sets} of negative trivia about a concept. We re-use the 200 concepts from before, but now produce top-5 negations for each, and show them to annotators at once. We compare the best version of our model (\uncommon{}\textsuperscript{S}) as the default, and the best of NegatER (NegatER-${\nabla}$), as well as Quasimodo\textsuperscript{neg} and GPT-3\textsuperscript{neg}. This results in a total of 2.4k annotations (200 concepts $\times$ 4 methods $\times$ 3 annotations). For every list of negations for a given concept, we ask the annotators whether the list is interesting, and allow again the same 3 options ``interesting'', ``slightly interesting'', and ``not interesting''. The Fleiss' kappa inter-annotator agreement is 0.24, i.e., fair. \uncommon{} leads with 49\% informativeness, followed by GPT-3\textsuperscript{neg} (40\%), NegatER (30\%), and finally Quasimodo\textsuperscript{neg} (23\%). An example is top negations about the concept \texttt{pancake}: While Quasimodo\textsuperscript{neg} and GPT-3\textsuperscript{neg} are low on plausibility, \texttt{$\neg$(pancake, is vegan)} and \texttt{$\neg$(pancake, is eaten)}, respectively, \uncommon{} offers the most plausible and informative negations e.g., \texttt{$\neg$(pancake, is crumbly)}.

\subsection{Extrinsic Evaluation II: KB Completion}

KB completion refers to the task of identifying novel \textit{positive} statements not yet in a KB. Recent works approach this as an LM-based true/false classification task on candidate statements~\cite{safavi-etal-2021-negater}. A crucial ingredient for this approach are negative examples for training the classifier, and this is where negation generation comes into play. Strong negative examples, i.e., nontrivial ones, can significantly benefit the classifier learning, and in turn, the KB completion accuracy. Following the setup of~\cite{safavi-etal-2021-negater}, we compare the impact of negations generated by \uncommon{} with that of COMET~\cite{comet} and NegatER\footnote{Based on data released at \url{https://github.com/tsafavi/NegatER/tree/master/configs/conceptnet/true-neg/}.}. 
We use the code by~\cite{safavi-etal-2021-negater} to train a BERT-based KB completion based on each of the three training datasets (100 randomized runs), and report the mean accuracy on the unseen test-set. The results are shown in Table~\ref{tab:kbcompletion}. \uncommon{} shows a statistically significant improvement over all methods with $\alpha < 0.01$. 

\begin{table}
\caption{KB completion evaluation.}\label{tab:kbcompletion}
\begin{tabular}{|l|c|}
\hline
\textbf{\cellcolor{gray!15}Negation Generator} & \multicolumn{1}{c|}{\textbf{\cellcolor{gray!15}Accuracy (\%)}} \\
\hline
CWA &  75.89\\
COMET & 79.06  \\
NegatER & 78.61  \\
\hdashline
\uncommon{} & \underline{79.56}  \\
\hline
\end{tabular}
\end{table}

\subsection{Extrinsic Evaluation III: Multiple-choice Question Answering}
Multiple-choice question answering (MCQA) is a common educational and entertainment evaluation setup. Humans approach MCQA often in two ways: (1) Via positive cues on what is the right answer, and (2) Via negative cues that eliminate incorrect answer options, thus narrowing down the set of possible answers. We next investigate to which degree negation generators can help in the second approach. We use the data from the CommonsenseQA task~\cite{talmor-etal-2019-commonsenseqa}. Examples are shown in Table~\ref{tab:eliminationexamples}. Every question comes with a question-concept (i.e., target concept) specifying the topic of the question. For example, the target concept  of ``\textit{Where can you store a pie?}'' is \texttt{pie}. The dataset contains 12k questions, each with only one correct answer. We manually sample 100 questions that: (1) Match concepts in the input KB (i.e., Ascent) and (2) Do \textit{not} require any additional condition or information (e.g., ``\textit{Where do people read newspapers \underline{while riding to work}?}''). We translate the questions to a KB-like triple-pattern. For instance, ``\textit{Where can you store a pie?}'' is mapped to \texttt{(pie, AtLocation, ?)}. For each question, the eliminator (e.g., \uncommon{}) crosses out the answers that \textit{match} a similar negation produced for the target concept (similarity is again measured using SBERT with threshold=0.7). The numerical results are shown in Table~\ref{tab:eliminationprocess} and examples in Table~\ref{tab:eliminationexamples}. A helpful elimination is an deletion of a \textit{wrong answer} and an unhelpful one is a deletion of a \textit{correct answer}. The CWA baseline eliminates most of the options since the absence of the statement is enough to merit a deletion. Besides CWA, the model with the highest number of helpful eliminations is \uncommon{} with 108, followed by NegatER with 35.

\begin{table}[!ht]
\caption{Eliminations for MCQA task.}\label{tab:eliminationprocess}
\begin{tabular}{|l|c|c|}
\hline
\textbf{\cellcolor{gray!15}Eliminator} &
\multicolumn{1}{c|}{\textbf{\cellcolor{gray!15}Helpful}} & \multicolumn{1}{c|}{\textbf{\cellcolor{gray!15}Unhelpful}} \\
\hline
CWA & 290 (72.5\%) & 72 (72.0\%)\\
Quasimodo\textsuperscript{neg}& 17 (4.3\%) & 1 (1.0\%) \\
NegatER &  35 (8.8\%) & 11 (11.0\%) \\
\hdashline
\uncommon{} &  108 (27\%) & 22 (22.0\%)\\
\hline
\end{tabular}
\end{table}

\section{Analysis}
\label{sec:ablation}

\noindent
\textbf{Ablation study. } In this study, our goal is to show the impact of every component in \uncommon{}. For instance, \textit{do the plausibility checks improve the correctness of the inferred negations?} and \textit{does the ranking improve the informativeness?} We run our method on the 200 concepts from Section~\ref{sec:intrinsic} and follow the same crowdsourcing setup for 4 different configurations of our method (4 configurations $\times$ 200 concepts $\times$ 2 negations $\times$ 3 annotators). The Fleiss' kappa inter-annotator agreement of this task is fair on both tasks, namely 0.33 on plausibility and 0.26 on informativeness. The results are shown in Table~\ref{tab:ablation}. One can see that without comparable concepts (instead random) to derive good thematic candidates from, the informativeness drops to almost half of the complete-configuration (i.e., \uncommon{}\textsuperscript{S}). This is different from the CWA baseline in Section~\ref{sec:experiments} in that we still scrutinize and rank the candidate set. The informativeness is also highly affected by the suspension of the ranking step (a decrease of 21\%). Moreover, holding off the plausibility checks shows an increase of 24\% in false negatives. 

\begin{table}
\caption{Ablation study results.}\label{tab:ablation}
\resizebox{\columnwidth}{!}{\begin{tabular}{|l|c|c|}
\hline
\textbf{\cellcolor{gray!15}Configuration} &  \textbf{\cellcolor{gray!15}False Negatives} & \textbf{\cellcolor{gray!15}Informativeness}\\
\hline
w/o comparable concepts  & 0.19 & 0.26\\
w/o quality checks & 0.28 & 0.22 \\
w/o plausibility checks & 0.49 & 0.38\\
w/o ranking & 0.39  & 0.29\\
{\cellcolor{gray!15}\textit{complete configuration}} & {\cellcolor{gray!15}\textit{0.25}} & {\cellcolor{gray!15}\textit{0.50}}\\
\hline
\end{tabular}}
\end{table}

\begin{figure}
\begin{subfigure}{\columnwidth}
\centering
\begin{tikzpicture}
\begin{axis}[
    height=3.5cm,
    width=.45\columnwidth,
    xlabel={$\gamma$},
    ylabel={\footnotesize informativeness},
    xmin=0, xmax=200,
    ymin=0, ymax=0.5,
    xtick={1,80,160},
    xticklabels={1,80,160},   
            ]
\addplot[smooth,mark=*,black] plot coordinates {
    (1,0.0)
    (20,0.3)
    (40,0.4)
    (60,0.3)
    (80,0.3)
    (100,0.3)
    (120,0.3)
    (140,0.3)
    (160,0.3)
    (180,0.3)
    (200,0.26)
};
\end{axis}
\end{tikzpicture}
\quad
\begin{tikzpicture}
\begin{axis}[
    height=3.5cm,
    width=.45\columnwidth,
    xlabel={$\lambda$},
    ylabel={\footnotesize informativeness},
    xmin=0, xmax=1,
    ymin=0, ymax=0.5,
    xtick={0,0.4,0.8,1},
    xticklabels={0,0.4,0.8,1},   
            ]
\addplot[smooth,mark=*,black] plot coordinates {
    (0,0.0)
    (0.1,0.0)
    (0.2,0.0)
    (0.3,0.0)
    (0.4,0.1)
    (0.5,0.2)
    (0.6,0.4)
    (0.7,0.4)
    (0.8,0.4)
    (0.9,0.3)
    (1.0,0.3)

};
\end{axis}
\end{tikzpicture}
\end{subfigure}
\bigskip
\begin{subfigure}[b]{\columnwidth}
\centering
\begin{tikzpicture}
\begin{axis}[
    height=3.5cm,
    width=.45\columnwidth,
    xlabel={$\tau$},
    ylabel={\footnotesize informativeness},
    xmin=0, xmax=100,
    ymin=0.3, ymax=0.4,
    xtick={1,40,80},
    xticklabels={1,40,80},   
    ytick={0,0.3,0.4}
            ]
\addplot[smooth,mark=*,black] plot coordinates {
    (1,0.34)
    (10,0.34)
    (20,0.35)
    (30,0.35)
    (40,0.35)
    (50,0.35)
    (60,0.34)
    (70,0.34)
    (80,0.33)
    (90,0.33)
    (100,0.33)

};
\end{axis}
\end{tikzpicture}
\quad
\begin{tikzpicture}
\begin{axis}[
    height=3.5cm,
    width=.45\columnwidth,
    xlabel={$\beta$},
    ylabel={\footnotesize informativeness},
    xmin=0, xmax=0.4,
    ymin=0, ymax=0.5,
    xtick={0,0.1,0.2,0.3,0.4},
    xticklabels={0,0.1,0.2,0.3,0.4},   
            ]
\addplot[smooth,mark=*,black] plot coordinates {
    (0,0)
    (0.01,0.29)
    (0.05,0.33)
    (0.1,0.25)
    (0.15,0.20)
    (0.2,0.18)
    (0.25,0.18)
    (0.3,0.18)
    (0.35,0.18)
    (0.4,0.18)
};
\end{axis}
\end{tikzpicture}
\end{subfigure}
\vspace{-1cm}
\caption{Hyperparameters Tuning.}
\label{fig:hyper}
\end{figure}
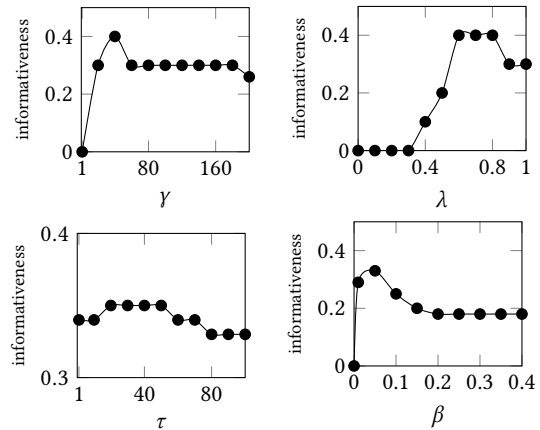

\noindent
\textbf{Hyperparameters Tuning. } Our methodology includes four main hyperparameters, namely $\gamma$ (number of comparable concepts), $\lambda$ (textual similarity threshold used in scrutinizing candidates and relaxed ranking), $\tau$ (rank threshold for LM), and $\beta$ (KB threshold for too-generic statements) . We experimented with different values for these parameters, and set them to their ideal values in Section~\ref{sec:experiments} as shown in Figure~\ref{fig:hyper}, namely $\gamma$ to 30, $\lambda$ to 0.7, $\tau$ to 50, and $\beta$ to 0.05.


\section{Related Work}
\noindent
\textbf{Commonsense knowledge bases. } Commonsense knowledge acquisition includes several large-scale projects. ConceptNet~\cite{conceptnet,speer-havasi-2012-representing}, the most prominent of these projects, was mainly constructed using human crowdsourcing. Similarly, ATOMIC~\cite{atomic} was also constructed using crowdsourcing, with its main focus on  collecting \textit{social} commonsense statements.  WebChild~\cite{webchild} uses handcrafted extraction patterns. TupleKB~\cite{TACL1064} and Quasimodo~\cite{quasimodo,insidequasimodo} rely on open information extraction~\cite{openie} followed by cleanup. Ascent~\cite{ascent,ascentpp} builds on these approaches by extending them to large-scale web text extraction, for better corroboration and higher recall.

\noindent
\textbf{Negation in knowledge bases. } ConceptNet~\cite{conceptnet} allows the expression of negative statements using 6 pre-defined negative relations. We use these statements in our recall evaluation.
The text-extracted Quasimodo~\cite{quasimodo} contains 350k negated statements (i.e., with negative \textit{polarity}), yet many have quality issues due to problems with the data source or extraction pipeline. We filter these negated statements from the full KB and use them as a baseline in our experiments (i.e., Quasimodo\textsuperscript{neg}). On actively collecting interesting negations, recently, an inference model has been proposed to build a knowledge graph~\cite{Pan2016}  with if-then commonsense contradictions~\cite{ANION}. Unlike our work, \cite{ANION} focus  on action-based statements and contradictions. For example, ``Wearing a mask is seen as responsible'' and ``Not wearing a mask is seen as carefree''. In terms of research problem and goal, the closest work to ours is NegatER~\cite{safavi-etal-2021-negater,negaterworkshop}. It proposes using LMs to discover meaningful negations. It fine-tunes the LM for statement truth classification and then uses similarity-based statement corruption to generate candidate negations. In the last step, these are ranked based on proximity to the LM's decision threshold, or a measure of model surprise. As our experiments show, although the methodology is interesting, 
the taxonomy-unaware corruptions of positive statements is not enough to obtain informative negations. Other approaches that target \textit{salient} negations in encyclopedic knowledge bases, such as Wikidata~\cite{WD} and Yago~\cite{YAGO}, include statistical inferences~\cite{arnaout2020enriching,ARNAOUT2021100661,10.1145/3442442.3452339,10.14778/3476311.3476350} and text extractions~\cite{AKB,arnaout2020enriching}. Yet text extraction is an inherently noisy process, and statistical inference over well-structured encyclopedic data does not carry over to verbose and non-canonicalized textual statements like in commonsense.

\noindent 
\textbf{Language Models. } In recent years, Language Models (LMs) showed their ability to store factual knowledge, learned from pre-training data~\cite{petroni-etal-2019-language,autoprompt}. Via LM-probing, one can predict missing tokens in a given claim, e.g. \textit{dogs can [MASK]} $\rightarrow$ \textit{walk, run, eat}. In addition, LMs can be trained to derive semantically meaningful sentence embeddings~\cite{reimers-2019-sentence-bert,gao-etal-2021-simcse}, which helps with the problem of detecting semantic similarity. However, LMs have also been repeatedly shown to struggle with explicit negation \cite{negatedlama,olmpics}. We make use of these models in order to scrutinize our candidate negations and make our rankings stronger via the relaxed sibling frequency. 

\section{Resources}
\label{sec:res}
We release a large dataset as a resource for further research: Up to\footnote{For some concepts, the final candidate set contains less than 1k candidates due to lack of enough siblings' positives and/or deletion during scrutiny steps} top-1k negations for all primary concepts from Ascent~\cite{ascentpp}, containing 6.2m negations\footnote{\url{https://www.mpi-inf.mpg.de/departments/databases-and-information-systems/research/yago-naga/commonsense/uncommonsense}}.

\section{Conclusion}
In this work, we presented the \uncommon{} framework for compiling informative negative statements about everyday concepts, by exploiting comparable concepts in commonsense knowledge bases. Our method outperforms baselines and state-of-the-art methods, on both informativeness and recall. Potential future directions include considering further types of negation~\cite{lobue-yates-2011-types}, e.g., conditioned and enriched with semantic facets \cite{ascent}, like ``\textit{female} lions do not have manes'', and exploring better sources for negative social knowledge \cite{atomic}, which comes with novel challenges due to a lack of previous work on taxonomic organization of activities.

\begin{acks}
Funded by the German Research Foundation (DFG: Deutsche Forsch\-ungsgemeinschaft) - Project 453095897 - ``Negative Knowledge at Web Scale''.
\end{acks}


\bibliographystyle{ACM-Reference-Format}
\balance
\bibliography{bibliography}
\end{document}